# Reinforcement Learning-based Defect Mitigation for Quality Assurance of Additive Manufacturing

Jihoon Chung[a], Bo Shen[b], Andrew Chung Chee Law[c], Zhenyu (James) Kong[a*]

[a]Grado Department of Industrial and Systems Engineering, Virginia Tech, Blacksburg, US

[b]Department of Mechanical and Industrial Engineering, New Jersey Institute of Technology, Newark, US

[c]IoTeX, Menlo Park, US

## Abstract

Additive Manufacturing (AM) is a powerful technology that produces complex 3D geometries using various materials in a layer-by-layer fashion. However, quality assurance is the main challenge in AM industry due to the possible time-varying processing conditions during AM process. Notably, new defects may occur during printing, which cannot be mitigated by offline analysis tools that focus on existing defects. This challenge motivates this work to develop online learning-based methods to deal with the new defects during printing. Since AM typically fabricates a small number of customized products, this paper aims to create an online learning-based strategy to mitigate the new defects in AM process while minimizing the number of samples needed. The proposed method is based on model-free Reinforcement Learning (RL). It is called Continual G-learning since it transfers several sources of prior knowledge to reduce the needed training samples in the AM process. Offline knowledge is obtained from literature, while online knowledge is learned during printing. The proposed method develops a new algorithm for learning the optimal defect mitigation strategies proven the best performance when utilizing both knowledge sources. Numerical and real-world case studies in a fused filament fabrication (FFF) platform are performed and demonstrate the effectiveness of the proposed method.

*Keywords:* Additive Manufacturing (AM), Model-free Reinforcement Learning (RL), G-Learning, Knowledge Transfer, Defect Mitigation, Fused Filament Fabrication (FFF)

## NOMENCLATURE

| | |
|---|---|
| $s_t$ | State ($s_t \in S$) at time $t$ (i.e., process parameter settings in AM process at time $t$). |
| $a_t$ | Action ($a_t \in A$) at time $t$ (i.e., an action that increases or decreases a process parameter level). |
| $r_t$ | Reward ($r_t \in R$) at time $t$ (i.e., reward provided based on the surface quality of printed parts). |
| $\alpha_t(s_t, a_t)$ | Learning rate at time $t$ when the state and action are $s_t$, $a_t$, respectively. |

* Corresponding author, Email: zkong@vt.edu

| | |
|---|---|
| $\gamma$ | Discounting factor. |
| $\pi(a_t\|s_t)$ | Probability of action $(a_t)$ given a state $(s_t)$, i.e., Policy. |
| $V(s_t)$ | The total expected reward of Reinforcement Learning starting from state $s_t$. |
| $Q(s_t, a_t)$ | State-action value of Q-Learning at time $t$. |
| $G(s_t, a_t)$ | State-action value of G-Learning at time $t$. |
| $GV(s_t)$ | The total expected reward of G-Learning starting from state $s_t$. |
| $CGV(s_t)$ | The total expected reward of Continual G-Learning starting from state $s_t$. |
| $CG(s_t, a_t)$ | State-action value of Continual G-Learning at time $t$. |
| $\rho(a\|s)$ | Prior policy of AM process in G-Learning. |
| $\rho_1(a\|s)$ | Offline prior policy of AM process in Continual G-Learning. |
| $\rho_2(a\|s)$ | Online prior policy of AM process in Continual G-Learning. |
| $\beta$ | Coefficient of prior policy in G-Learning. |
| $\beta_1, \beta_2$ | Coefficient of offline $(\beta_1)$ and online $(\beta_2)$ prior policy in Continual G-Learning. |

## 1. Introduction

Additive Manufacturing (AM), also known as "3D printing," makes a three-dimensional shape from a digital model. The Fused Filament Fabrication (FFF) process is one of the most widely used AM technologies attributed to its low cost and material flexibility [1, 2]. FFF uses a movable head that heats a thermoplastic filament to melting temperatures and then extrudes onto a substrate. This extruded material solidifies and subsequently bonds to the previous layers [3, 4]. During this repeated solidifying and bonding phase, some defects such as voids, over-fill, and under-fill may occur [5, 6]. These defects can cause a severe discrepancy in geometrical tolerance, loss of internal structure precision, and poor surface quality of AM products [7, 8]. Aiming to mitigate defects of AM products, many research efforts have been reported in the literature, such as post-processing [9, 10], design of experiments (DOE) [11], and mathematical optimization methods [12].

However, these methods mentioned above are primarily offline analysis tools, and they cannot identify and correct defects during printing. For example, Figure 1 shows the limitation of DOE in the AM process. Figure 1 (a) illustrates the CAD model for a printed part using the predetermined offline optimal process parameters based on the DOE. Figure 1 (b) and (c) show the surface quality of the 3rd and 30th layers of the part, respectively. The surface quality of the 30th layer shows under-fill defects due to the accumulation of uncertainties from the complex process such as machine vibration, ambient temperature, and humidity [13].

Therefore, online adjustments of the process parameters are necessary. To fulfill this need, [13] proposed a method based on closed-loop quality control for the FFF process using a PID controller to mitigate defects via online process parameter adjustments. However, this method can only handle the defects identified and trained by the controller beforehand. It cannot deal with new defects during printing that was not recognized by the system.

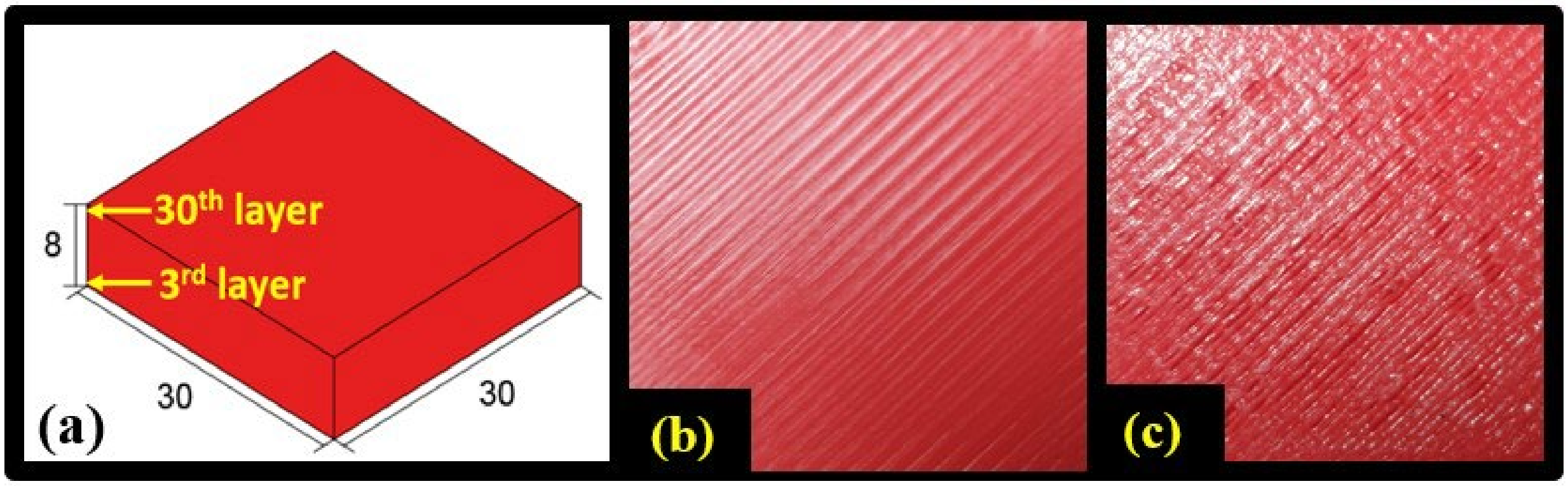

*Figure 1. (a) CAD model for a printed part with units of mm; (b) Normal printing in 3rd layer based on predetermined optimal parameter based on the DOE; (c) Under-fill defects observed in 30th layer during printing.*

This limitation motivates the online learning-based method to deal with new defects that did not occur during the previous printing. Specifically, the online learning-based method needs to learn the new process parameter adjustments to mitigate the new defects. Compared to conventional manufacturing that produces a large number of parts, AM processes usually fabricate a small number of customized products [14]. Thus, the training samples are relatively limited when used for modeling purposes. Several existing methods address this issue by transfer learning (TL) in shape deviation analysis [15, 16]. They built a statistical model first based on an AM process (e.g., shape deviation model in Sec. 2.2) and then transferred it to a new process. However, they did not study the online defect mitigation problem.

This paper aims to develop an online learning-based method to mitigate the new defects in AM process with a limited number of samples. The baseline of the proposed method is a model-free Reinforcement Learning (RL) method, namely, G-learning [17], that does not require any model of AM process, which is challenging to develop due to its complexity and uncertainties. G-Learning can incorporate prior knowledge to reduce training samples of model-free RL [17]. However, it can transfer only one source of prior knowledge, but multiple sources may exist in the AM process. For example, offline knowledge can be obtained from literature or previous experiments, while online knowledge can be learned during printing. The utilization of multiple sources of offline and online knowledge is beneficial for quickly learning how to mitigate new defects. To transfer both sources of prior knowledge to the current AM process, the proposed method, namely, Continual G-Learning, is developed in this paper. Specifically, the method provides an algorithm that learns the optimal defect mitigation strategy while utilizing both sources of prior knowledge. To demonstrate the effectiveness of the proposed method, a real-world case study in the FFF platform is conducted. To the best of our knowledge, this is the first work that uses a model-free RL-based method with various sources of knowledge for defect mitigation in AM processes.

The rest of this paper is organized as follows. A brief review of related research work is provided in Sec. 2 to identify the research gap. The overall research framework is introduced in Sec. 3. The proposed research methodology is presented in Sec. 4, followed by the case studies to validate the proposed method in Sec. 5 and Sec. 6. Finally, the conclusions and future work are discussed in Sec. 7.

## 2. Review of Related Work

The existing studies on surface quality assurance in the FFF processes are reviewed in Sec. 2.1. Then, the literature related to TL in AM processes is provided in Sec. 2.2. Afterward, the research gaps of the current work are identified in Sec. 2.3.

### 2.1 Existing Surface Quality Assurance for the FFF Process

FFF processes are vulnerable to surface defects because the thermoplastic properties of filaments that determine the ability to create a bond between layers and solidify the extruded filament are sensitive to the environment [18]. There were several proposed solutions using post-processing, DOE, mathematical optimization methods, and closed-loop control to deal with surface defects. [19] proved that significant improvements on the surface finish of acrylonitrile butadiene styrene (ABS) parts could be achieved using the chemical post-processing treatment. [9] used an acetone vapor bath for post-process smoothing to reduce the surface roughness and reach a maximum 95% reduction in surface roughness. Using the DOE method, [20] determined the effect of layer thickness and deposition speed on the surface roughness of the FFF process. [21] used factorial design to improve the surface roughness of ABS 400 polymer materials in the FFF process. [11] studied the effect of process variables on surface texture parameters to predict the surface roughness by the Taguchi method. In addition, there were several research efforts to deal with quality issues in the FFF process by optimization of mathematical models. [22] applied a genetic algorithm to determine the optimum part deposition orientation to improve the surface quality by measuring the arithmetic mean of surface roughness. [12] used a particle swarm optimization algorithm to obtain a target surface quality. Recently, [13] developed a closed-loop controller for the FFF process. It consists of the image-based process monitoring [23] and PID controller for defect mitigation.

### 2.2 TL in the Additive Manufacturing Process

TL can overcome the challenge of a limited number of samples in AM processes [24, 25]. Some previous research uses TL for shape deviation modeling in AM processes. [26] proposed a statistical TL that is related to shape deviation modeling. This method modeled the dimensional error of a product by a parameter-based TL approach. Specifically, it transfers the statistical model for shape-independent error to a new part so that the shape deviation model for a new product can be built with limited samples. [15, 16] used TL to deal with shape deviation models in different manufacturing conditions and processes. To avoid re-collecting the entire training data in a new condition, they transfer the deviation model across other AM processes to build a new deviation model.

### 2.3 Research Gap Analysis

The work summarized in Sec. 2.1 contributed to the defect identification and mitigation for online process monitoring and control of AM processes. But they assumed that the defects that occurred in the process had been identified in advance and thus are not suitable for handling new defects during printing. On the other hand, the learning-based method requires mitigating new defects with limited samples in AM process. Research efforts in Sec. 2.2 used TL to deal with limited samples for shape deviation modeling in AM processes. However, these efforts do not deal with defect mitigation. In addition, it is a challenging task to build the statistical model in defect mitigation because of the complex relationship between process parameters and surface quality. To overcome the above limitations, this paper proposed a TL-based method, namely, Continual G-Learning, that can detect and mitigate new defects during printing with limited samples. The proposed model-free RL method can reduce training samples by transferring offline and online prior knowledge to the current AM process. The proposed method also provides theoretical proof that the method learns the optimal defect mitigation strategy when utilizing both offline and online prior knowledge.

## 3. Research Framework

The overall research framework of online learning-based defect mitigation in AM process is provided in Figure 2. The framework iterates the following three steps:

- Step 1: Collect surface images in the FFF 3D printing process;
- Step 2: Detect surface defects using an image-based classifier (e.g., one-class support vector machine (SVM) [27]);
- Step 3: Mitigate the defects by learning how to adjust process parameters (e.g., printing speed, layer height, flow rate multiplier, etc.).

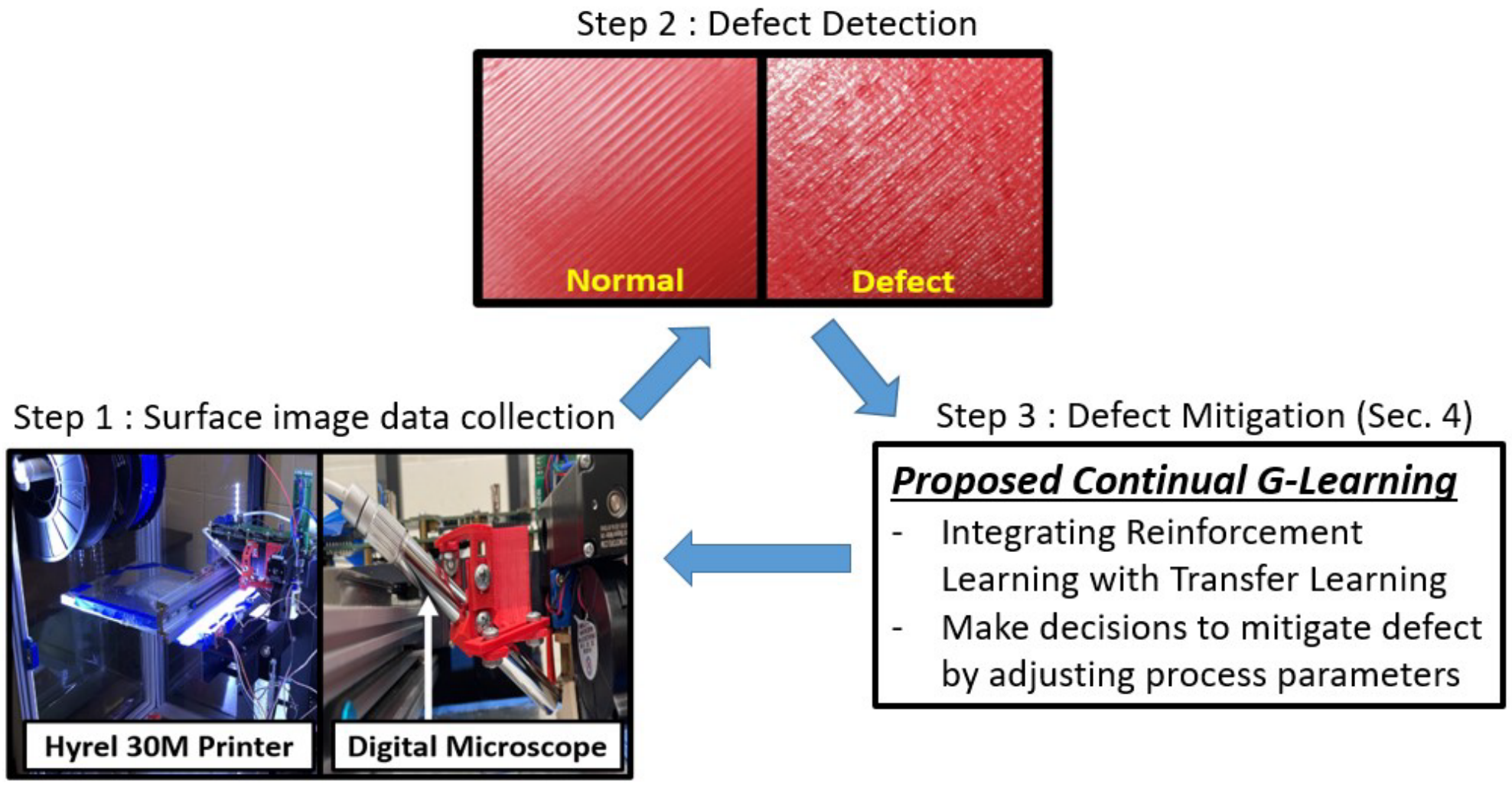

*Figure 2. The proposed research framework.*

This study uses a Hyrel System 30M 3D printer (Hyrel 3D, Norcross, GA, USA), an FFF machine, as shown in Figure 2. In Step 1, a digital microscope collects online surface images, and ABS is the printing material. An image-based classifier is implemented to classify the surface images and detect defects in Step 2. Once a defect is detected, it is mitigated in Step 3 through our proposed method, namely, Continual G-Learning, to adjust process parameters. Continual G-Learning is based on RL and utilizes several sources of prior knowledge to learn the optimal decisions to mitigate the defects accurately and quickly. Step 3 in the framework with the proposed method is described in detail in Sec 4.

## 4. Integration of RL and TL for AM Defect Mitigation

This section presents the proposed research methodology, which integrates RL and TL for AM defect mitigation. RL can be used as a tool for learning decisions to mitigate a new defect in AM process, described in Sec. 4.1. G-Learning [17] is a representative TL method in RL applied to reduce the number of training samples in Sec. 4.1. Finally, a new approach, namely, Continual G-learning, is proposed in Sec. 4.2 to integrate the two types of prior knowledge (i.e., offline and online) in the AM process to further reduce the needed training samples.

### 4.1 Application of RL and TL for Defect Mitigation in AM Processes

Once a defect is detected and identified as a new one during the printing process, there are no available solutions to mitigate it. The control system needs to identify a possible solution quickly to mitigate the new defect. The key here is the decision on the change of process parameters currently being used, by which the new defect will be mitigated.

RL is applied here for such purpose, based on the Markov decision process (MDP). MDP is a 4-tuple $(S, A, P, R)$, where $S$ is collection of states that describe the feasible processes parameter setting (e.g., printing speed, layer height, flow rate multiplier, etc.), and $s_t \in S$ is the state at time *t*; $A$ is an action set that consists of increasing or decreasing the level of process parameters (i.e., parameter adjustments), and $a_t \in A$ is an action performs at time $t$; $P$ is a transition probability between states, with $P(s_{t+1}|s_t, a_t)$ denoting the probability of transition to state $s_{t+1}$ from state $s_t$ when action $a_t$ is taken; $R(s_t, a_t, s_{t+1})$ represents reward function depending on the states and actions, which is determined based on the surface quality improvement or deterioration in the AM process, and $r_t$ (i.e., $R(s_t, a_t, s_{t+1})$) is the reward incurred at time $t$ in the process. The surface defects cause a deficiency in mechanical properties of the final product, such as density, tensile strength, and compressive strength [28]. Therefore, the defects need to be mitigated promptly to prevent quality losses in the AM process. To achieve this, RL learns the decisions that can mitigate the defect as soon as possible when it occurs. In other words, the goal of RL is to learn an optimal decision (i.e., policy used in the following context) that maximizes the total expectation of reward ($r_t$) incurred in the AM process based on the surface quality. The total expected reward in RL, namely, $V(s)$, is formulated as a state value function as follows [29]:

$$V(s) = \sum_{t\geq 0} \gamma^t \mathbb{E}_\pi[r_t | s_0 = s], \tag{1}$$

where $\gamma$ and $\pi$ denote discount factor and policy, respectively, and the discount factor is in the range of $0 \leq \gamma \leq 1$, which specifies the weights of future rewards. The policy, $\pi$, is a probability distribution of actions in each state. Specifically, a policy is a probability distribution of process parameter adjustments in the current parameter setting in AM process. $r_t$ is provided as a positive value when the defect is successfully mitigated from the AM process; otherwise, $r_t$ is zero. RL learns the policy that maximizes Eq. (1), by which the shortest sequence of process parameter adjustments will be generated to mitigate the defect. If the transition probability $P(s_{t+1}|s_t, a_t)$ and reward function $R(s_t, a_t, s_{t+1})$ is known, the optimal policy that maximizes Eq. (1) can be obtained by model-based RL [30]. However, it is challenging to estimate the accurate transition probability and reward function in AM process because of the high complexity and uncertainties of the process.

Model-free RL [31] learns optimal policy without estimating the transition probability and the reward function of AM process. Instead, it learns an optimal policy directly based on the samples $(s_t, a_t, r_t, s_{t+1})$ that obtained from the interaction with AM process. Q-Learning is a representative model-free RL [31]. The objective of Q-Learning is to learn policy $\pi$ that maximizes the total expected reward, which is quantified as state-action value function $Q$ as follows:

$$Q(s, a) = \sum_{t\geq 0} \gamma^t \mathbb{E}_\pi[r_t | s_0 = s, a_0 = a]. \tag{2}$$

Q-Learning updates $Q$ value from time $t$ to $t + 1$ based on Eq. (3) as follows [31, 32]:

$$Q_{t+1}(s_t, a_t) = (1 - \alpha_t(s_t, a_t))Q_t(s_t, a_t) + \alpha_t(s_t, a_t)\left(r_t + \gamma \max_a Q_t(s_{t+1}, a)\right), \tag{3}$$

where $0 \leq \alpha_t(s_t, a_t) \leq 1$ is the learning rate in time $t$ in state $s_t$ with action $a_t$. Eq. (3) shows that the $Q$ value is updated with the reward ($r_t$) measured by the surface quality and the maximum expected reward starting with subsequent process parameter settings ($\max_a Q_t(s_{t+1}, a)$). Since Q-Learning is a model-free RL method, the transition probability is not required in Eq. (3). In addition, $r_t$ which is obtained from AM process is directly used for updating $Q$ value, instead of using the estimated reward function. The optimal policy $\pi$ in state $s$ is the parameter adjustment (i.e., action) that results in maximum $Q$ value described as follows:

$$\pi(a|s) = \arg\max_{a} Q(s, a).$$

Q-Learning assumes there is no prior knowledge about defect mitigation, so it learns the policy from scratch. However, in actual AM processes, some general knowledge can be obtained from our previous experience or literature, such as the melting temperature range or printing speed of each material in AM process for the target surface quality. Utilizing this prior knowledge will improve the effectiveness and efficiency of the learning process for AM. As one of the most representative TL approaches in RL, G-Learning [17] can utilize the prior knowledge, which is applied as the baseline for this study. The prior knowledge is used as a prior policy in G-Learning. Denote $\rho(a|s)$ and $\pi(a|s)$ as a prior policy and a policy to be learned, respectively. The divergence between $\rho(a|s)$ and $\pi(a|s)$ is defined as the information cost as follows [17]:

$$g^{\pi}(s, a) = \log\frac{\pi(a|s)}{\rho(a|s)}. \tag{4}$$

The expectation of the information cost represents the Kullback-Leibler divergence (KL-divergence) between both policies as follows:

$$\mathbb{E}_{\pi}[g^{\pi}(s, a)|s] = D_{KL}[\pi||\rho]. \tag{5}$$

Eq. (5) represents the divergence between the policy to be learned and prior knowledge in AM process. By considering both reward incurred in the AM process and the information cost, the total expected reward of G-Learning is represented as state value function $GV(s)$ as follows [17]:

$$GV(s) = \sum_{t\geq 0} \gamma^{t} \mathbb{E}_{\pi}\left[r_t + \frac{1}{\beta} g^{\pi}(s_t, a_t) \middle| s_0 = s\right], \tag{6}$$

where $\beta < 0$ is a coefficient of information cost. By maximizing Eq. (6), G-Learning learns the optimal policy that maximizes the reward of the AM process while penalizing the policy that diverges from prior knowledge of the AM process.

### 4.2 The Proposed Continual G-learning using More Prior Information

In this section, a new approach named Continual G-learning that integrates both offline and online prior policies is developed. Compared to G-Learning, which only uses one source of prior knowledge, the proposed method aims to transfer prior knowledge from two sources. Specifically, it can transfer both offline knowledge and online knowledge simultaneously. Offline prior knowledge is the knowledge that can be acquired before printing, such as knowledge that can be obtained from literature or previous experiments. In contrast, online knowledge is the knowledge that is learned during printing. When G-Learning in Sec. 4.1 completes to learn the optimal policy by utilizing the offline prior policy in the AM process, an image-based classifier of the proposed framework in Sec. 3 provides a positive reward. However, when the properties of the process, such as the geometry of the part being printed changes at different layers, the classifier would provide a zero reward in the same process parameters since the parameters are not optimal in a new geometry. This provides a signal for the transition from G-Learning to Continual G-Learning in the proposed framework. Then, the optimal policy learned from G-Learning becomes the online prior knowledge in the proposed method, allowing the proposed method to utilize both prior knowledge sources.

Let $\rho_1(a|s), \rho_2(a|s)$ as offline and online prior policies in AM process, respectively. Instead of Eq. (4), the information cost of $\pi(a|s)$ is defined as

$$g_1^\pi(s,a) + g_2^\pi(s,a) = \log\frac{\pi(a|s)}{\rho_1(a|s)} + \log\frac{\pi(a|s)}{\rho_2(a|s)}, \tag{7}$$

where $g_1^\pi(s,a) = \log\frac{\pi(a|s)}{\rho_1(a|s)}$, $g_2^\pi(s,a) = \log\frac{\pi(a|s)}{\rho_2(a|s)}$. The expectation of Eq. (7) provides KL divergence between prior policies and $\pi(a|s)$:

$$\mathbb{E}_\pi[g_1^\pi(s,a) + g_2^\pi(s,a)|s] = D_{KL}[\pi||\rho_1] + D_{KL}[\pi||\rho_2].$$

Considering both information cost from offline and online prior policies and the reward earned from the AM process, the total expected reward in Continual G-Learning is defined as its state value function as follows:

$$CGV(s) = \Sigma_{t\geq 0}\gamma^t \mathbb{E}_\pi\left[r_t + \frac{1}{\beta_1} g_1^\pi(s_t,a_t) + \frac{1}{\beta_2} g_2^\pi(s_t,a_t)\middle| s_0 = s\right], \tag{8}$$

where $\beta_1 < 0$ and $\beta_2 < 0$ are the coefficients of information cost of each prior knowledge, respectively. To derive the optimal policy of the proposed Continual G-Learning, state-action value function of the proposed method is required and defined as,

$$CG(s,a) = \sum_{t\geq 0} \gamma^t \mathbb{E}_\pi \left[ r_t + \frac{\gamma}{\beta_1} g_1^\pi(s_{t+1}, a_{t+1}) + \frac{\gamma}{\beta_2} g_2^\pi(s_{t+1}, a_{t+1}) \middle| s_0 = s, a_0 = a \right]. \tag{9}$$

By plugging Eq. (9) into Eq. (8), the state value function in Eq. (8) can be reformulated as follows:

$$CGV(s) = \sum_a \pi(a|s) \left[ \frac{1}{\beta_1} log \frac{\pi(a|s)}{\rho_1(a|s)} + \frac{1}{\beta_2} log \frac{\pi(a|s)}{\rho_2(a|s)} + CG(s,a) \right]. \tag{10}$$

By maximizing Eq. (10) with constraint $\Sigma_a \pi(a|s) = 1$, Continual G-Learning learns the policy that maximizes the reward incurred from the AM process by penalizing the deviations from both prior policies (i.e., knowledge). It represents that both offline and online prior knowledge guide the learning direction of $\pi(a|s)$ in the learning procedure. Therefore, the knowledge aids the proposed method to learn how to mitigate defects quickly. Based on the method of Largrange multipliers [33], the policy in Eq. (10) is derived as

$$\pi(a|s) = \frac{\rho_1(a|s)^{\frac{\beta_2}{\beta_1+\beta_2}} \rho_2(a|s)^{\frac{\beta_1}{\beta_1+\beta_2}} e^{-CG(s,a)\frac{\beta_1\beta_2}{\beta_1+\beta_2}}}{\Sigma_a \rho_1(a|s)^{\frac{\beta_2}{\beta_1+\beta_2}} \rho_2(a|s)^{\frac{\beta_1}{\beta_1+\beta_2}} e^{-CG(s,a)\frac{\beta_1\beta_2}{\beta_1+\beta_2}}}. \tag{11}$$

Compared to Q-Learning starts from the policy that selects the action randomly (i.e., random policy), $\pi(a|s)$ in Eq. (11) is initialized with various sources of prior knowledge about defect mitigation since state-action value, namely, $CG(s,a)$ is initialized as zero. The proposed Continual G-Learning provides an update rule of state-action value ($CG(s,a)$) in Eq. (12). Based on this rule, the value converges to optimal state-action value proven theoretically by Theorem 1 that is described later in this section, and the optimal state-action value leads to optimal policy $(a|s)$ in Eq. (11). The update rule of state-action value from time $t$ to $t+1$ can be written as follows:

$$CG_{t+1}(\mathrm{s_t}, a_t) = \left(1 - \alpha_t(s_t, a_t)\right) CG_t(s_t, a_t) + \alpha_t(s_t, a_t)(r_t - \gamma \frac{\beta_1 + \beta_2}{\beta_1 \beta_2} \times \log(\sum_{a'} [\rho_1(a'|s_{t+1})^{\frac{\beta_2}{\beta_1+\beta_2}} \rho_2(a'|s_{t+1})^{\frac{\beta_1}{\beta_1+\beta_2}} e^{-CG_t(s_{t+1}, a')\frac{\beta_1\beta_2}{\beta_1+\beta_2}}]), \quad (12)$$

where the learning rate $\alpha_t(s_t, a_t)$ is defined as $n_t(s_t, a_t)^{-w}$. $n_t(s_t, a_t)$ is the number of times that the pair $(s_t, a_t)$ is visited until time $t$, and $w \in (0.5, 1]$ is learning rate hyperparameter. Eq. (12) represents that the state-action value in the proposed method is updated by the reward $(r_t)$ and the subsequent process parameter settings $(\mathrm{s}_{t+1})$ that obtained by process parameter adjustment $(a_t)$. Specifically, the value is updated by both the reward $(r_t)$ in time $t$, and the maximum expected reward starting from $\mathrm{s}_{t+1}$ and follows policy $\pi(a|s)$ in Eq. (11) $(-\gamma \frac{\beta_1+\beta_2}{\beta_1\beta_2} \times \log(\sum_{a'} [\rho_1(a'|s_{t+1})^{\frac{\beta_2}{\beta_1+\beta_2}} \rho_2(a'|s_{t+1})^{\frac{\beta_1}{\beta_1+\beta_2}} e^{-CG_t(s_{t+1}, a')\frac{\beta_1\beta_2}{\beta_1+\beta_2}}])$. The algorithm of the proposed Continual G-Learning is summarized in Figure 3. Starting from an initial process parameter settings in AM process, Continual G-Learning adjusts process parameters based on the policy in Eq. (11) (line 6 in Figure 3). Based on the parameter adjustments, the method reaches subsequent process parameter settings and receives the reward based on the surface quality in the AM process (line 7 in Figure 3). Then, the state-action value in Continual G-learning is updated by Eq. (12) (line 10 in Figure 3). The value is used to update policy in Eq. (11). This algorithm iterates until the method achieves the optimal process parameter settings (i.e., terminal state) to mitigate the defects or reach the maximum number of iterations. The entire procedure is named an episode. The episode (from line 3 to line 13 in Figure 3) is repeated until the state-action value converges to optimal, eventually learning the shortest number of parameter adjustments to mitigate the defects.

**Algorithm 1** Continual G-Learning

Require: State $S$, Action $A$, Coefficient $\beta_1, \beta_2$, Discounting factor $\gamma \in [0,1]$,
Maximum number of iterations in an episode: $iter_{max}$,
Learning rate hyperparameter $w$.

Initialize $CG_0(s,a) = n_0(s,a)^{-w} = 0, \forall s, a$.
While $CG_t(s,a), \forall s, a$, are not converged, do
  Start from the initial state $s_t \in S, iter = 0$.
  While state $s_t \in S$ is not terminal state or $iter \leq iter_{max}$
    Calculate $\pi(a|s_t)$ as in Eq. (11)
    Choose $a_t \in A$ using policy derived from $\pi(a|s_t)$
    Obtain $s_{t+1}, r_t$
    $n_t(s_t, a_t) = n_t(s_t, a_t) + 1$
    $\alpha_t(s_t, a_t) = n_t(s_t, a_t)^{-w}$
    Calculate $CG_{t+1}(s_t, a_t)$ based on Eq. (12)
    $t \leftarrow t + 1$
    $iter \leftarrow iter + 1$
  Return $CG_t(s,a), \forall s, a$.

*Figure 3. Algorithm of Continual G-Learning.*

The theoretical convergence of state-action value to the optimal state-action value based on Algorithm 1 is provided as follows. To begin with, necessary definitions and results to build the convergence of Algorithm 1 are introduced. Let $CG^*(s_t, a_t)$ be the optimal state-action value of state $s_t$ and action $a_t$. Subtracting the quantity $CG^*(s_t, a_t)$ from both sides of Eq. (12) and letting

$$\Delta_t(s,a) = CG_t(s,a) - CG^*(s,a),$$

yields

$$\Delta_{t+1}(s_t, a_t) = \big(1 - \alpha_t(s_t, a_t)\big)\Delta_t(s_t, a_t) + \alpha_t(s_t, a_t) F_t(s_t, a_t),$$

where $F_t(s_t, a_t) = \left(r_t - \gamma \frac{(\beta_1+\beta_2)}{(\beta_1\beta_2)} \log\left(\sum_{a'}\left[\rho_1(a'|s_{t+1})^{\frac{\beta_2}{(\beta_1+\beta_2)}} \rho_2(a'|s_{t+1})^{\frac{\beta_1}{(\beta_1+\beta_2)}} e^{-CG_t(s_{t+1},a')\frac{\beta_1\beta_2}{(\beta_1+\beta_2)}}\right]\right)\right) - CG^*(s_t, a_t)$. $\Delta_{t+1}(s,a)$ represents the difference between a state-action value in time $t+1$ and optimal state-action value in state *s* and action $a$. To prove the convergence of state-action value in the proposed method, it is sufficient to prove Theorem 2 in [34] that a random iterative process $\Delta_{t+1}(s_t, a_t)$ converges to zero $w.p.1$ under the following assumptions:

1. $0 \leq \alpha_t(s,a) \leq 1, \sum_t \alpha_t(s,a) = \infty$ and $\sum_t \alpha_t(s,a)^2 < \infty, \forall s \in S, a \in A$;

2. $\|\mathrm{E}[F_t(s,a)|U_t]\|_\infty \le \gamma\|\Delta_t\|_\infty$, with $\gamma < 1$;

3. $\mathrm{var}[F_t(s,a)|U_t] \le K(1+\|\Delta_t\|_\infty^2)$, for $K > 0$,

where $U_t = \{\Delta_t, \Delta_{t-1}, \dots \Delta_0, F_{t-1}, \dots F_0\}$. $\|\cdot\|_\infty$ refers to supremum norm, and $K$ is a constant.

For any policy $\pi$, operator $B^\pi[CG(s,a)]$ is defined as follows:

$$B^\pi[CG(s,a)] = k^\pi(s,a) + \gamma \sum_{s',a'} p(s'|s,a)\pi(a'|s')CG(s',a'), \tag{13}$$

where

$$k^\pi(s,a) = E_p[r(s,a,s')] + \gamma \sum_{s',a'} p(s'|s,a)\pi(a'|s')[\frac{1}{\beta_1}\log\frac{\pi(a'|s')}{\rho_1(a'|s')} + \frac{1}{\beta_2}\log\frac{\pi(a'|s')}{\rho_2(a'|s')}].$$

To prove the convergence of our proposed algorithm, Lemma 1 is used to prove the second assumption above.

**Lemma 1.** For any policy $\pi$, the operator $B^\pi[CG(s,a)]$ is a contraction under the supremum norm over *s*, *a*. That is, for any $CG1(s,a)$ and $CG2(s,a)$, it follows

$$\|B^\pi[CG1(s,a)] - B^\pi[CG2(s,a)]\|_\infty \le \gamma\|CG1(s,a) - CG2(s,a)\|_\infty .$$

*Proof.* The proof is provided in Appendix A.

Based on Lemma 1, the main theorem about the convergence of proposed Continual G-Learning can be stated as follows.

**Theorem 1.** Supposed that $0 < \rho_{min} \le \rho_1(a|s), \rho_2(a|s) \le \rho_{max} < 1$ for all $(s,a)$ and $\alpha_t(s_t,a_t) = n_t(s_t,a_t)^{-w}$ for $w \in (0.5, 1]$. Then, three assumptions in Theorem 2 in [34] are satisfied where $\gamma$ is the discount factor and $K = \max\left\{K' + \max_{s\in S, a\in A} CG^*(s,a)^2, 64\gamma^2\right\}$. $K'$ is defined as $2\mathbb{E}[R(s,a,s') - CG^*(s,a)]^2 + 4\gamma^2\left(\frac{\beta_1\beta_2}{\beta_1+\beta_2}\right)^2[2(\log|A|)^2 + 4(\log\rho_{min})^2 + 4(\log\rho_{max})^2]$. Therefore,

$$\lim_{t\to\infty} \Delta_t(s,a) \xrightarrow{w.p.1} 0.$$

*Proof.* The proof is provided in Appendix B.

Based on **Theorem 1**, Algorithm 1 is guaranteed to converge to optimal state-action value. Since optimal state-action value provides optimal policy from Eq. (11), the proposed Continual G-Learning learns the shortest parameter adjustments sequences to mitigate the defect by transferring offline and online prior knowledge in AM process.

## 5. Numerical Case Study

A numerical case study is performed in this section to evaluate the proposed Continual G-Learning performance. The Grid world-based simulation [35] is used for illustration of the effectiveness of the proposed Continual G-Learning. Random-Policy selecting the action randomly without any learning process and Q-Learning [31] and G-Learning [17] in Sec. 4.1 are selected as benchmark methods to compare with the proposed Continual G-Learning. To compare the performance of our proposed method with the benchmark methods, a total number of actions to complete a certain number of episodes is used as performance metrics, which is widely adopted in the RL algorithms as performance measures [36, 37]. The smaller number of actions to complete the episodes represents that method learns the optimal policy more quickly.

### 5.1 Description of Grid World

In the grid world, the unavailable squares are occupied by walls, shown in black in Figure 4. An episode starts from an initial state ($s_t \in S$ in Sec. 4) and terminates when reaching a goal state in Figure 4 or reaching the maximum number of iterations (i.e., $iter_{max}$). Each method repeats a number of episodes to learn the optimal policy. At each square, a method moves one square in any of the four directions, namely, left, right, up, and down ($a_t \in A$ in Sec. 4). If a move is blocked by the wall or the edge of the board, it stays in the same place.

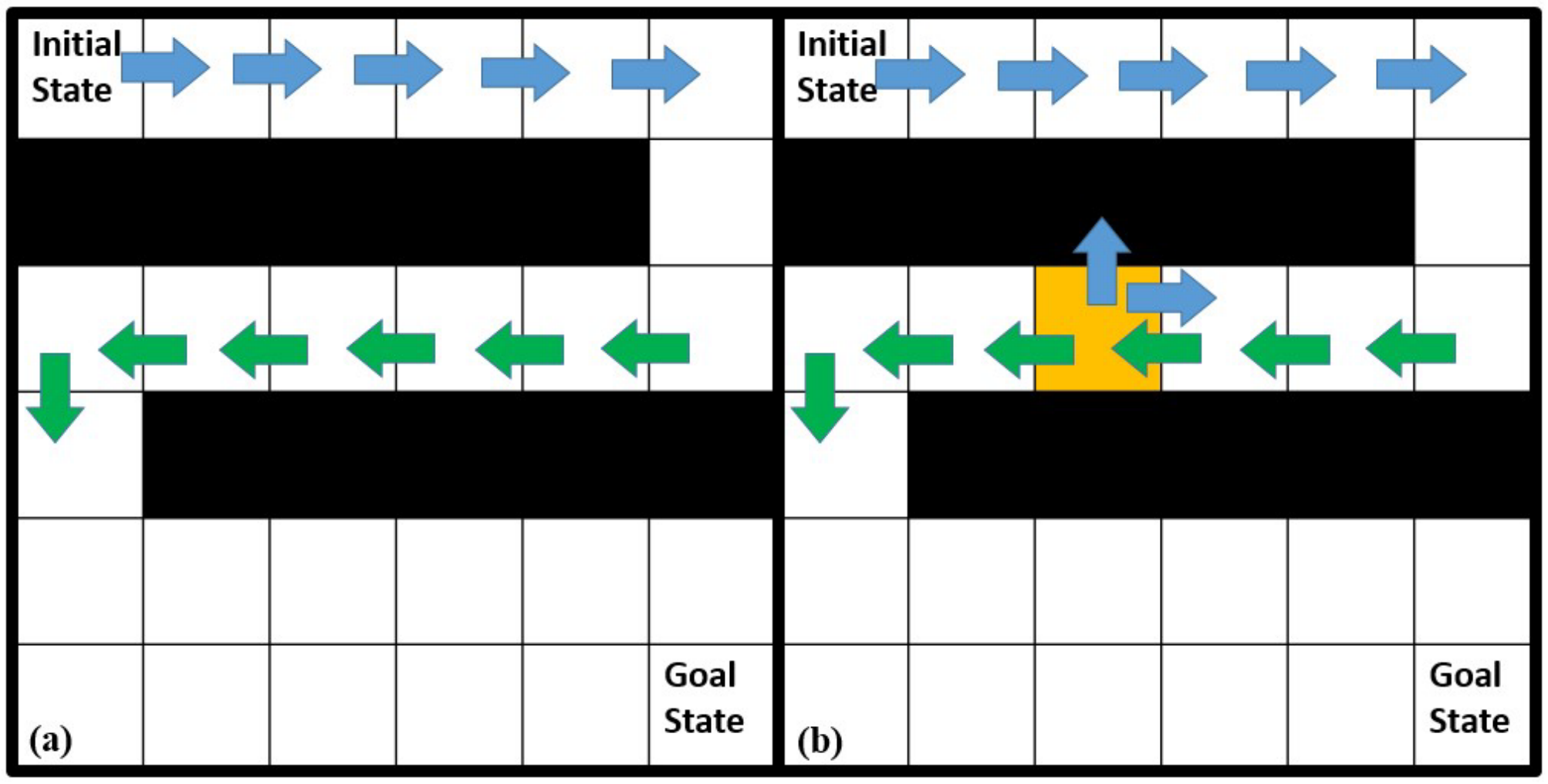

*Figure 4. Grid world domain with prior policies. (a) Case with all the prior policies following the optimal policy; (b) Case with not all the prior policies following the optimal policy. Arrows represent actions with the highest probability in a prior policy in each state.*

Arrows in Figure 4 denote actions with the highest probability in a prior policy in each state. For example, in Figure 4 (a) and (b), blue arrows point the right direction in an initial state. It implies the prior policy of moving to the right has a higher probability (i.e., 0.9) than the probability of the other three directions (i.e., 0.03 in each direction) in the initial state. The policy is called an informative prior policy. On the other hand, if all directions of prior policy have the same probability as 0.25, it is called random prior policy. The prior policy in G-Learning ($\rho(a|s)$ in Eq. (4)) uses the informative policy in states with blue arrows and random policy in the remaining states. The first prior policy in Continual G-Learning ($\rho_1(a|s)$ in Eq. (7)) is defined as the same policy as the prior policy in G-Learning. The second prior policy in the proposed method ($\rho_2(a|s)$ in Eq. (7)) consists of informative policy in states with green arrows and random policy in the remaining states. Detailed information of prior policy is illustrated in Appendix C. In this study, two cases are investigated, namely, (a) all the prior policies correspond to optimal policies (i.e., all the arrows in Figure 4 (a) have the same direction with optimal policy), (b) there exists a state with a prior policy that disagrees with the optimal policy that can hinder the learning process (e.g., blue arrows in a state with orange color in Figure 4 (b)).

**5.2 Performance Evaluation**

The average number of actions to complete 100 episodes in 50 replications is used as a performance evaluation measure. Hyperparameters used in this study are provided in Table 1. The learning rate hyperparameter ($w$) is chosen as 0.6 to meet the first condition of Theorem 2 in [34]. The reward is provided as 1 when a method reaches the goal state. Otherwise, the reward is assigned as zero. The discount factor ($\gamma$) is selected as 0.9 which is a typical value used in RL when the reward provided in the terminal state is larger than other states like in this simulation study [38]. $\beta$ and $\beta_1, \beta_2$ are the negative values which are the coefficients of information cost of each prior policy in G-Learning and Continual G-Learning, respectively. When the coefficients are small, both methods learn the policy that approaches prior knowledge since information cost in Eqs. (6) and (8) are dominant. As shown in Figure 4, there exist prior policies that correspond to the optimal policies. However, many prior policies, such as random policies, disagree with the optimal policies. Therefore, the coefficients are determined by tuning, and grid search [39] is used for tuning in this case study. To reduce the computational burden in a grid search, $\beta_1$ and $\beta_2$ in the proposed method are assumed to be equal. The coefficients are searched at intervals of 100 in the range of -500 to -3000.

*Table 1. Hyperparameters in the numerical case study.*

| Hyperparameters | Value |
|---|---|
| $w$ | 0.6 |
| $r$ (reach to goal state / otherwise) | 1 / 0 |
| $\gamma$ | 0.9 |
| $\beta = \beta_1 = \beta_2$ | $-2 \times 10^3$ |
| $iter_{max}$ | 1000 |

Figure 5 (a) and (b) show the performance evaluation of all methods in Cases (a) and (b), respectively. Continual G-Learning has the smallest number of actions to complete episodes in both Cases (a) and (b). After several episodes, the proposed method converges to the optimal policy, which is the validation of Theorem 1 in Sec. 4.2. Performance of G-Learning is significantly degraded in Case (b) compared to Case (a) since there exist prior policies that can hinder the learning process. The performance is similar to that of Q-Learning, which does not utilize any prior knowledge. However, Continual G-Learning overcomes this challenge by utilizing an additional source of prior policy that corresponds to optimal policy (i.e., green arrow in the state with orange color in Figure 4 (b)). Therefore, Continual G-Learning has a similar performance in both Cases (a) and (b). Random-Policy shows the worst performance among the benchmark methods. As shown in Figure 5, the performance of Random-Policy does not have improvements over episodes since this approach is not learning-based.

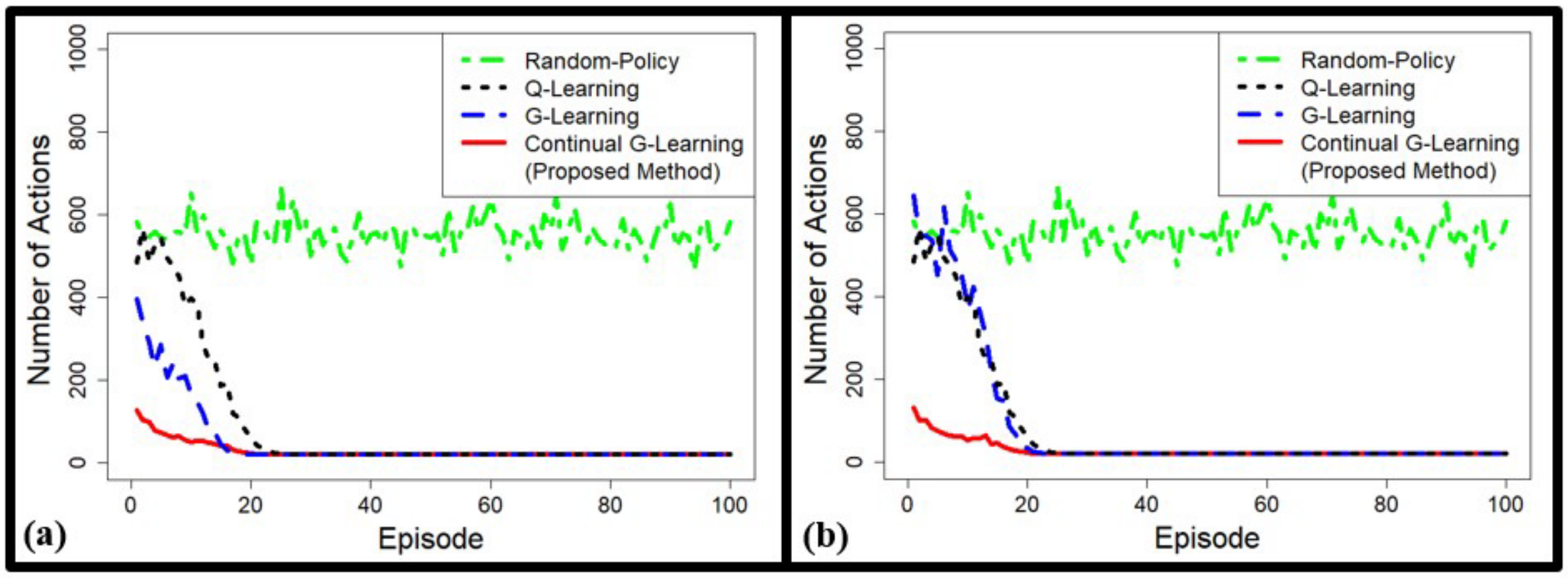

*Figure 5. The number of actions to complete episodes when state size is* 6× 6 *with the (a) prior policies corresponding to optimal policies and (b) prior policies do not follow optimal policies, respectively.*

To demonstrate the performance of the proposed method with different sizes of state space, the general rule is designed to provide blue and green arrows in Cases (a) and (b) when state size is $n \times n$. Let $(i, j)$ as an index of the state. For both Cases (a) and (b), an initial state is $(1,1)$, and a goal state is $(n, n)$. Unavailable states are $(2, j)$, where $j = 1, .., n-1$, and $(4, j)$, where $j = 2, \ldots, n$. A blue rightward pointing arrow is provided to states $(i, 1)$, where $i = 1, .., n-1$. A green leftward pointing arrow is used to states $(3, j)$, where $j = 2, .., n$. A green downward pointing arrow is allowed to a state (3,1). In addition, blue upward and leftward pointing arrows are provided to a state (3,4) in Case (b). The arrows represent the prior policy that can hinder the learning process.

Table 2 summarizes the performance evaluations of all methods in different sizes of state. Since Random-Policy and Q-Learning do not utilize any prior policies, the results of Cases (a) and (b) are the same. Compared to the result in Case (a), the performance of G-Learning in Case (b) in all different state sizes is significantly deteriorated because of the prior policy, which hinders the learning process. However, the proposed Continual G-Learning shows similar results in both cases in every size of states, and it shows the best performance compared to benchmark methods.

*Table 2. The average number of actions to complete 100 episodes in 50 replications by varying the size of the state in Cases (a) and (b). RP, QL, GL, and CGL denote Random-Policy, Q-Learning, G-Learning, and Continual G-Learning, respectively.*

| | Case (a) | | | | Case (b) | | | |
|---|---|---|---|---|---|---|---|---|
| State size | RP | QL | GL | CGL | RP | QL | GL | CGL |
| $6 \times 6$ | 55323.8 | 8459.0 | 4766.7 | **2786.4** | 55323.8 | 8459.0 | 8633.5 | **2832.8** |
| $7 \times 7$ | 65441.6 | 14163.7 | 7128.5 | **3693.8** | 65441.6 | 14163.7 | 13607.2 | **3750.4** |
| $8 \times 8$ | 74680.9 | 22752.7 | 10162.1 | **4686.3** | 74680.9 | 22752.7 | 21952.9 | **4690.6** |
| $9 \times 9$ | 82648.3 | 34426.3 | 15018.4 | **6104.7** | 82648.3 | 34426.3 | 30385.1 | **6193.8** |
| $10 \times 10$ | 88494.9 | 50000.2 | 20524.1 | **7823.5** | 88494.9 | 50000.2 | 42515.1 | **7829.6** |

## 6. Real-World Case Study

This section shows a real-world case study based on the FFF platform to test our proposed Continual G-Learning. The part has two different geometries that are printed sequentially (Figure 6 (a)):

1) Geometry 1: a cuboid with size 30 mm × 30 mm × 6 mm on the bottom, and
2) Geometry 2: a cuboid with size 15 mm × 15 mm × 18 mm on the top.

For clarity, the bottom and top parts of the print are denoted as Geometry 1 and Geometry 2, respectively. This case study aims to learn the optimal process parameter adjustments (i.e., policy) in both geometries to meet the target surface quality by mitigating defects, as shown in Figure 6 (b). Specifically, compared to the benchmark methods, the proposed method quickly learns the shortest sequence of decisions from current process parameters to the optimal process parameters. This real-world case study uses the same benchmark methods except for Random-Policy because of its poor performance that was validated in Sec. 5. The performance evaluation criterion used in Sec. 5 is utilized in this case study.

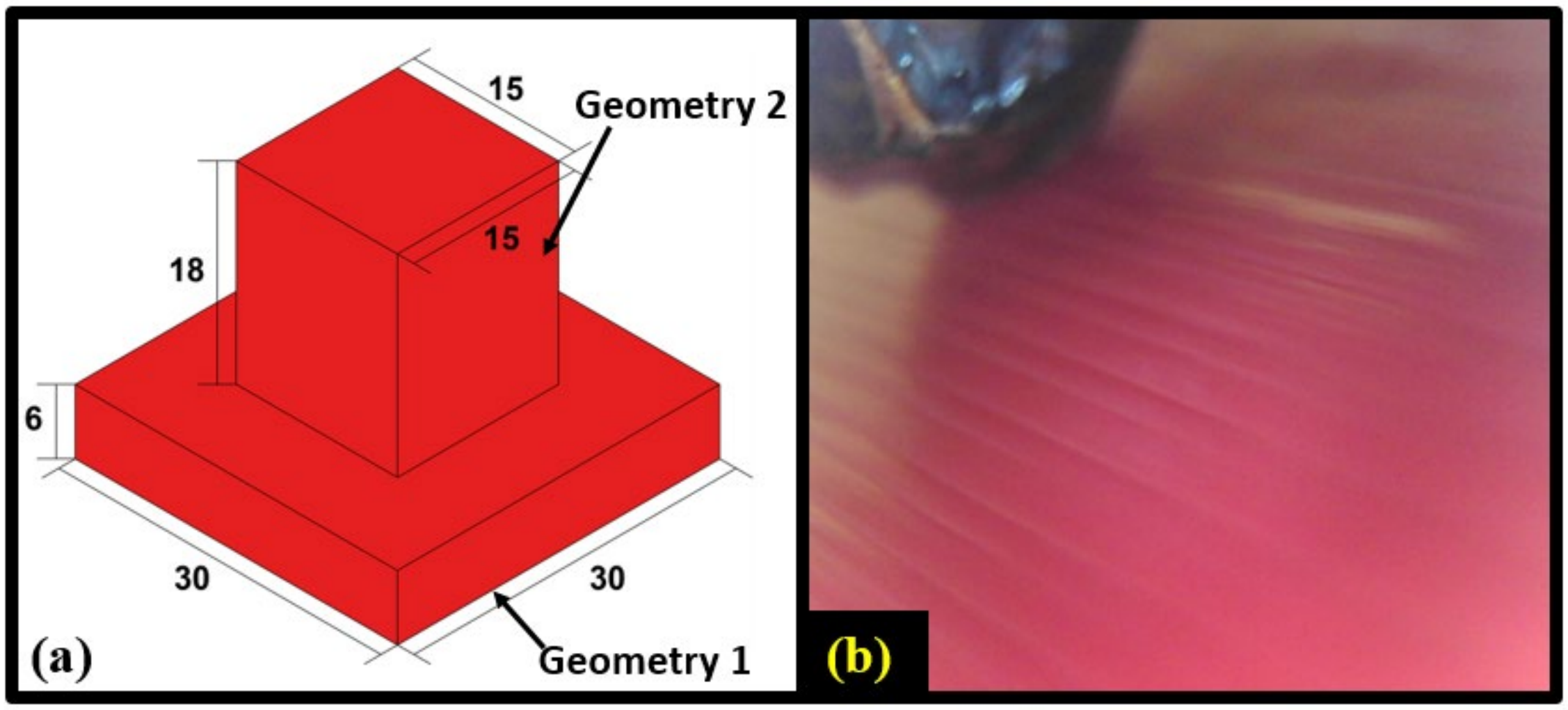

*Figure 6. (a) CAD model for print in case studies with units of mm; (b) Target surface quality.*

Sec. 6.1 describes the AM experimental platform used in this case study. In Sec. 6.2, state ($s_t \in S$ in Sec. 4), action ($a_t \in A$ in Sec. 4), and reward ($r_t$ in Sec. 4) are defined for our AM application. Description of experiments and performance evaluation are provided in Secs. 6.3 and 6.4, respectively. Finally, the printed part with optimal parameter setting is illustrated in Sec 6.5.

### 6.1 Experimental Platform

A Hyrel System 30M 3D printer (Hyrel 3D, Norcross, GA, USA) equipped with a 0.5mm extruder nozzle is used for this case study. Figure 7 (a) shows a front view of the printer. ABS (Hatchbox, Pomona, CA, USA) is used for printing with a diameter of 1.75mm. In every episode, the temperature of the extruder starts from 245℃, which is in the printing temperature range for ABS [40]. An Opti-Tekscope Digital Microscope Camera (Opti-Tekscope, Chandler, AZ, USA) is utilized for online image acquisition of surface quality, as shown in Figure 7 (b). The camera is mounted near the extruder to collect images of the surface that are being printed. A cooling fan is installed next to the extruder to cool down the surface of the printed part. Figure 7 (c) shows an open communication-based software controller. It allows the proposed Continual G-Learning to adjust the process parameters (in the form of G-code) during printing. A virtual serial port (RS-232) is used to communicate between the 3D printer controller and the external program that runs the proposed Continual G-learning algorithm. Defect detection and mitigation are executed based on the surface images acquired from the camera and the proposed method.

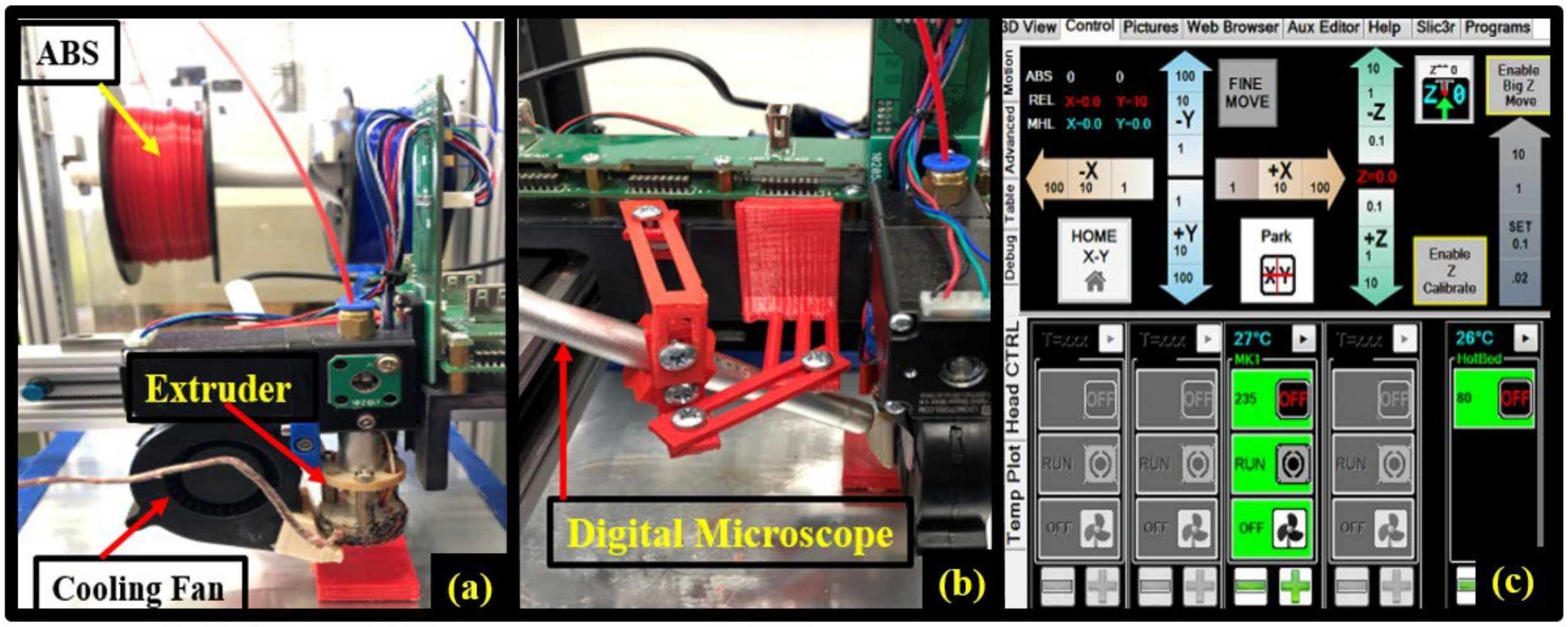

*Figure 7. (a) Front view of Hyrel system 30M; (b) Digital Microscope Camera; (c) Software controller.*

### 6.2 Description of State, Action, and Reward in the FFF Applications

Three process parameters, namely, flow rate multiplier, printing speed, and cooling fan, are adjusted in this case study. The printing speed denotes the speed of the extruder head in (mm/min). The flow rate multiplier indicates how much plastic the printer is to extrude. For example, a flow rate with a multiplier of 1.0 indicates the extruder would deposit material at normal flow ($mm^3$/s), while a multiplier of 0.8 or 1.2 indicates the extruder would deposit 20% less or 20% more material. The cooling fan can be controlled in terms of the operation of the fan. Each process parameter in this case study has two levels, as shown in Table 3.

*Table 3. Level representation of each process parameter.*

| Parameters \ Levels | 1 | 2 |
|---|---|---|
| Flow rate multiplier | 0.4 | 1.0 |
| Printing speed (mm/min) | 7500 | 2500 |
| Cooling fan | Off | On |

The action is defined as tuning the level of a single process parameter from the current setting. The state is defined as the combination of levels of each parameter in Table 3. The reward is provided based on the surface quality resulting from the current action. To measure surface quality, the offline trained one-class support vector machine (SVM) [27] is utilized as an image-based classifier. The classifier is trained with the features of target surface quality images (Figure 6 (b)) extracted by a pooling layer of pre-trained ResNet, which is a standard feature extraction method in many vision applications [41, 42]. This offline trained classifier, named a target classifier, predicts the quality of the surface image captured online by the digital microscope as target surface quality or an anomaly that is not. To collect the surface image data, a window-based approach is used. A window size of 21 is utilized. Namely, the 21 consecutive surface images are captured by a digital microscope with a sampling frequency of 5Hz. Surface quality is determined by voting from the classification results of 21 images from the target classifier. The window size and sampling frequency are determined to provide a robust classification result to unintended noise in the process, and they are validated from the previous printing. The reward is provided as a positive value if the surface quality is classified as the target surface, otherwise provides zero.

Hyperparameters used in the real-world case study are presented in Table 4, and they are selected for the same reasons provided in Sec. 5.

*Table 4. Hyperparameters in the real-world case study.*

| Hyperparameters | Value |
|---|---|
| $w$ | 0.6 |
| $r$ (reach to goal state / otherwise) | 1 / 0 |
| $\gamma$ | 0.9 |
| $\beta = \beta_1 = \beta_2$ | $-7 \times 10^2$ |
| $iter_{max}$ | 50 |

## 6.3 Description of Experiments

Three experiments are performed in this case study. Since the printed part consists of two different geometries (i.e., Geometry 1 and Geometry 2, respectively), each experiment consists of combinations of two methods summarized in Table 5. Additionally, the prior knowledge that transferred in each geometry is illustrated in Table 5. For offline knowledge, observation from [13] that surface defects are minimized when flow rate multiplier approaches one is used. The knowledge encourages methods with offline prior knowledge to select flow rate multiplier as one in high probability. Online knowledge is the optimal policy learned from Geometry 1. Detailed information of prior policy is illustrated in Appendix C. Experiment 1 uses G-Learning in Geometry 1 with offline knowledge. Then, it uses the proposed Continual G-Learning in Geometry 2 by transferring both offline and online knowledge. Experiments 2 and 3 start to print Geometry 1 without offline prior knowledge. Therefore, they use Q-Learning in Geometry 1. In Geometry 2, experiment 2 transfers offline prior knowledge by G-Learning, and experiment 3 uses Q-Learning.

*Table 5. Experiments description based on which prior knowledge (in the parenthesis) is used in each geometry.*

| | **Geometry 1** | **Geometry 2** |
|---|---|---|
| Experiment 1 | G-Learning (Offline) | Continual G-Learning (Offline, Online) |
| Experiment 2 | Q-Learning (None) | G-Learning (Offline) |
| Experiment 3 | Q-Learning (None) | Q-Learning (None) |

## 6.4 Performance Evaluation

In Geometry 1, the cooling fan is excluded from the process parameters since the temperature of the extruder (245℃) is in the range of printing temperatures of ABS (220℃~270℃) [40]. The initial state of Geometry 1 is the state with a flow rate multiplier of 0.4 and a printing speed of 7500 mm/min that causes surface defects, named as Defect 1. Since the Defect 1 is classified as an anomaly from the target classifier, it is identified as new defect. Several images of Defect 1 are collected to train the one-class SVM, denoted as Defect 1 classifier. The classifier is used to identify new defects in further printing. After the training the classifier, the proposed Contin al G-Learning starts to learn the parameter adjustments to mitigiate the new defect. The episode in the proposed method starts from the initial state and terminates when it reaches optimal parameter setting that produces the target surface quality. Figure 8 (a) shows the performance evaluations in Geometry 1. Based on offline knowledge, G-Learning learns the sequences of decisions from the initial state to the optimal parameter setting that flow rate multiplier of 1.0 and printing speed of 2500 mm/min in the 1st episode. It performs the same parameter adjustments in the 2nd and 3rd episodes. Therefore, the number of layers in Geometry 1 for each episode is constant, as shown in Figure 8 (b). The Q-Learning that does not use prior knowledge needs several more actions in the 1st episode to learn the optimal policy.

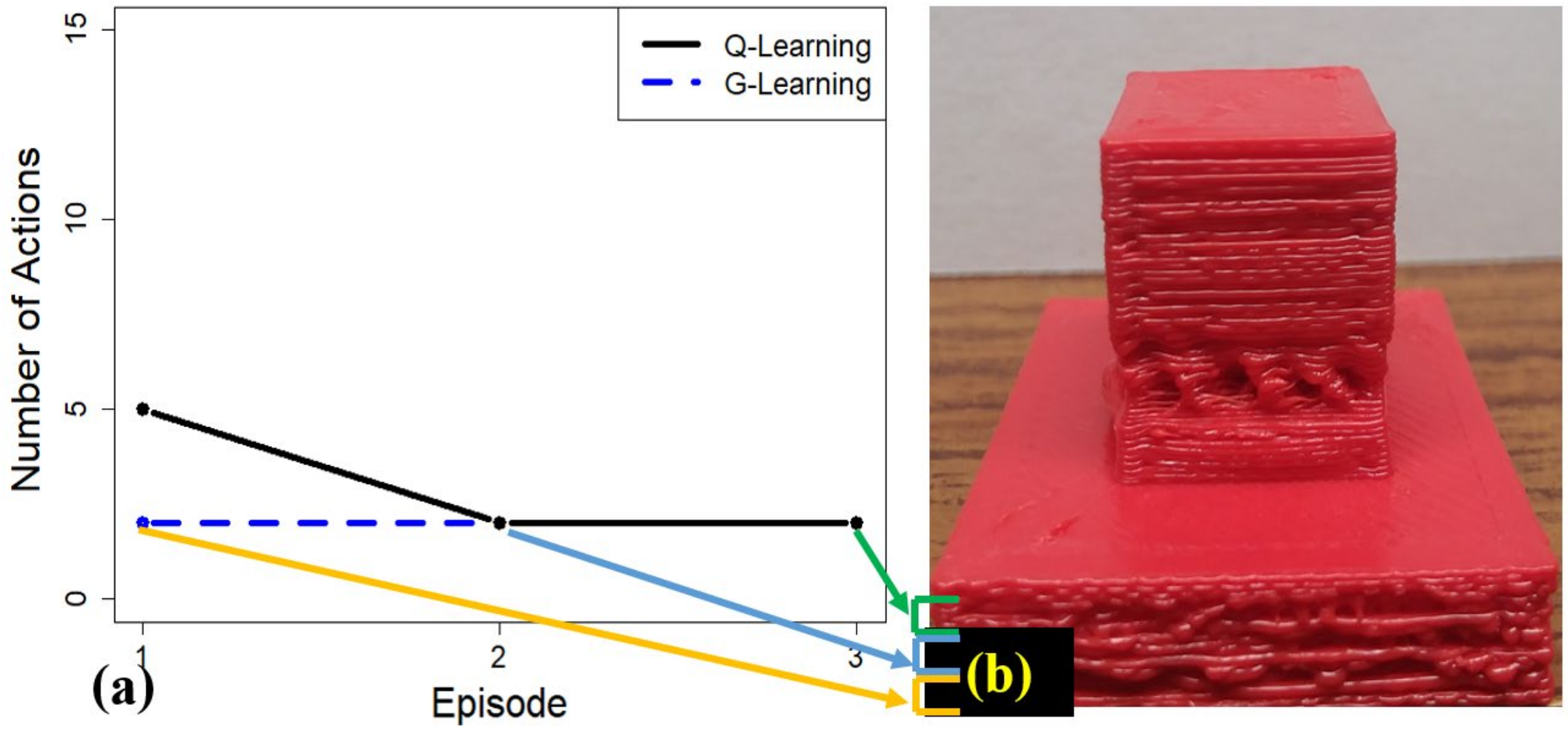

*Figure 8. (a) Number of actions needed to reach the target surface quality in Geometry 1 using G-Learning; (b) Corresponding layers of Geometry 1 in each episode using G-Learning.*

Due to the printing sequence, the episode's initial state in Geometry 2 is the optimal process parameter setting learned from Geometry 1. However, the surface quality from Figure 9 (b) shows that the optimal setting in Geometry 1 is no longer optimal in Geometry 2 anymore, and the surface quality is classified as an anomaly from the target classifier representing it as the defect. In addition, the surface is classified as an anomaly in the Defect 1 classifier, indicating that the surface is a new defect. It implies the learned process parameter adjustments from Geometry 1 is not suitable to mitigate this defect, and the new optimal process parameters need to be learned. This new defect in Geometry 2 is caused by insufficient time for layers to solidify before reheating due to the small size of Geometry 2 [43]. Therefore, the cooling fan becomes one of the process parameters that need to be adjusted in Geometry 2. If the temperature of the extruder is below 210℃, the fan is turned off irrespective of parameter setting to avoid the nozzle from being clogged [40].

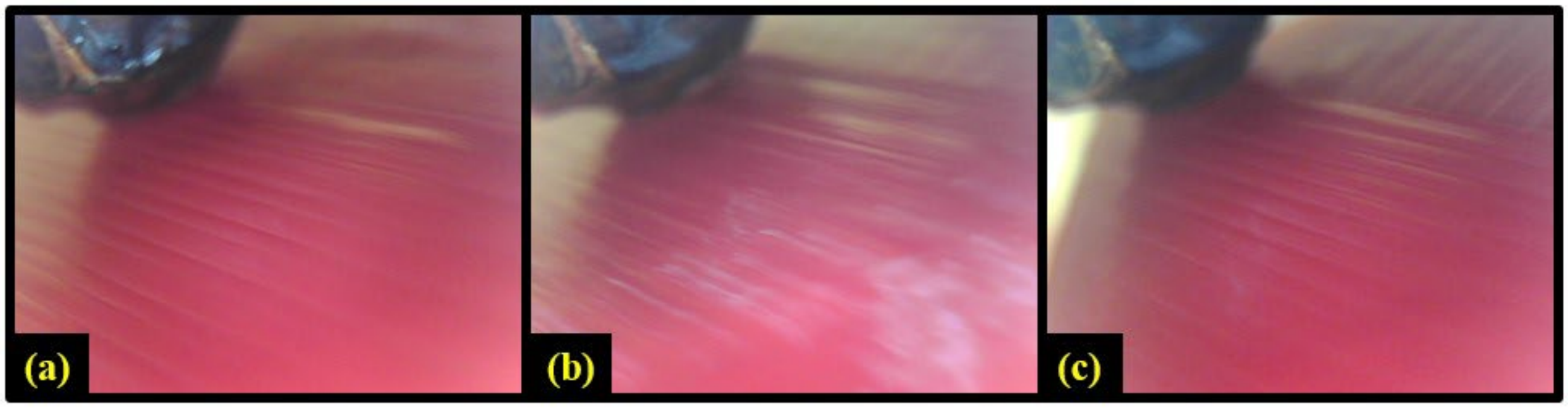

*Figure 9. (a) Surface quality in Geometry 1 with optimal parameter setting for Geometry 1 (target surface quality); (b) Surface quality in Geometry 2 with optimal parameter setting for Geometry 1; (c) Surface quality Geometry 2 with optimal parameter setting for Geometry 2.*

Figure 10 (a) shows the performance evaluations in Geometry 2. Continual G-Learning needs fewer actions to learn the optimal parameter adjustments than other methods by using offline and online prior knowledge. The knowledge encourages the flow rate and printing speed to set 1.0 and 2500 mm/min, respectively. The proposed method learns the optimal parameter adjustments from the 3rd episode based on the knowledge. The optimal process parameter setting in Geometry 2 is a flow rate multiplier of 1.0, printing speed of 2500 mm/min, and turning on the cooling fan. Figure 9 (c) shows the surface quality in Geometry 2 with the optimal parameter setting. It offers a similar surface quality to Figure 9 (a), the target surface quality collected in Geometry 1. Figure 10 (b) shows that the number of layers in Geometry 2 that need to be completed is reduced over episodes. It shows the proposed method learns the optimal policy as the episode increases.

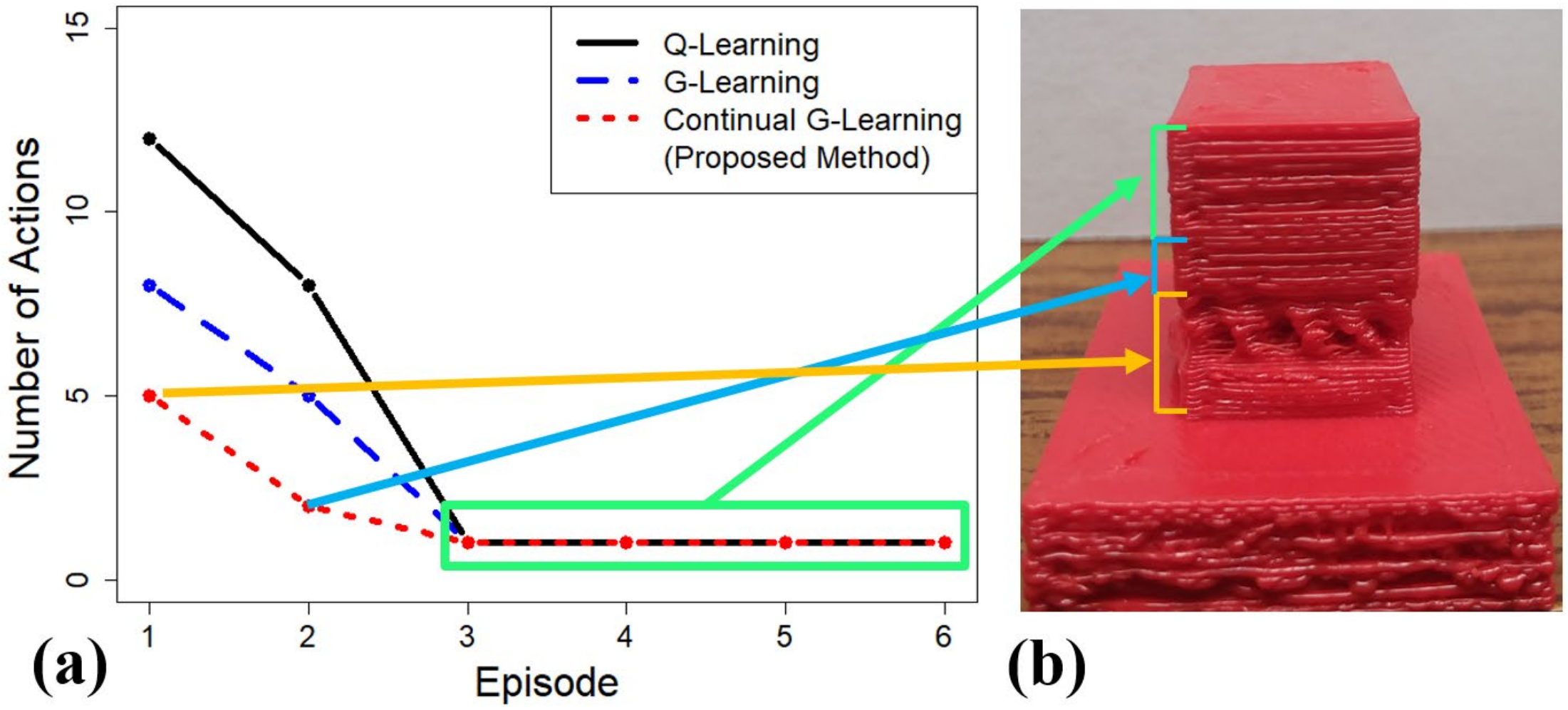

*Figure 10. (a) Number of actions needed to reach the target surface quality in Geometry 2 using Continual G-Learning; (b) Corresponding layers of Geometry 2 in each episode using Continual G-Learning.*

Experiment 1 with the proposed Continual G-Learning needs the least number of actions to learn the optimal parameter adjustments to meet the target surface quality in both geometries by transferring both sources of prior knowledge. Table 6 shows the number of actions needed in Geometries 1 and 2 to complete 3 and 6 episodes, respectively. Table 7 illustrates the optimal process parameters for defect mitigation learned from each geometry.

*Table 6. The number of actions (in the parenthesis) required to complete three episodes in Geometry 1 and six episodes in Geometry 2 for each experiment.*

| | **Geometry 1** | **Geometry 2** | **Total** |
|---|---|---|---|
| Experiment 1 | G-Learning (**6**) | Continual G-Learning (**11**) | **17** |
| Experiment 2 | Q-Learning (9) | G-Learning (17) | 26 |
| Experiment 3 | Q-Learning (9) | Q-Learning (24) | 33 |

*Table 7. Optimal process parameters for defect mitigation in each geometry.*

| | **Flow rate multiplier** | **Printing speed (mm/min)** | **Cooling fan** |
|---|---|---|---|
| Geometry 1 | 1.0 | 2500 | Off |
| Geometry 2 | 1.0 | 2500 | On |

### 6.5 Verification of the Learned Optimal Parameter Setting

Figure 11 shows a newly printed part using the learned optimal parameter settings in Geometries 1 and 2, listed in Table 7. The optimal parameter setting of Geometry 1 is flow rate multiplier with one and printing speed with 2500 mm/min. The optimal parameter setting for Geometry 2 is the same as Geometry 1 while turning on the cooling fan. Figure 11 shows defect-free print by optimal parameter settings in both geometries.

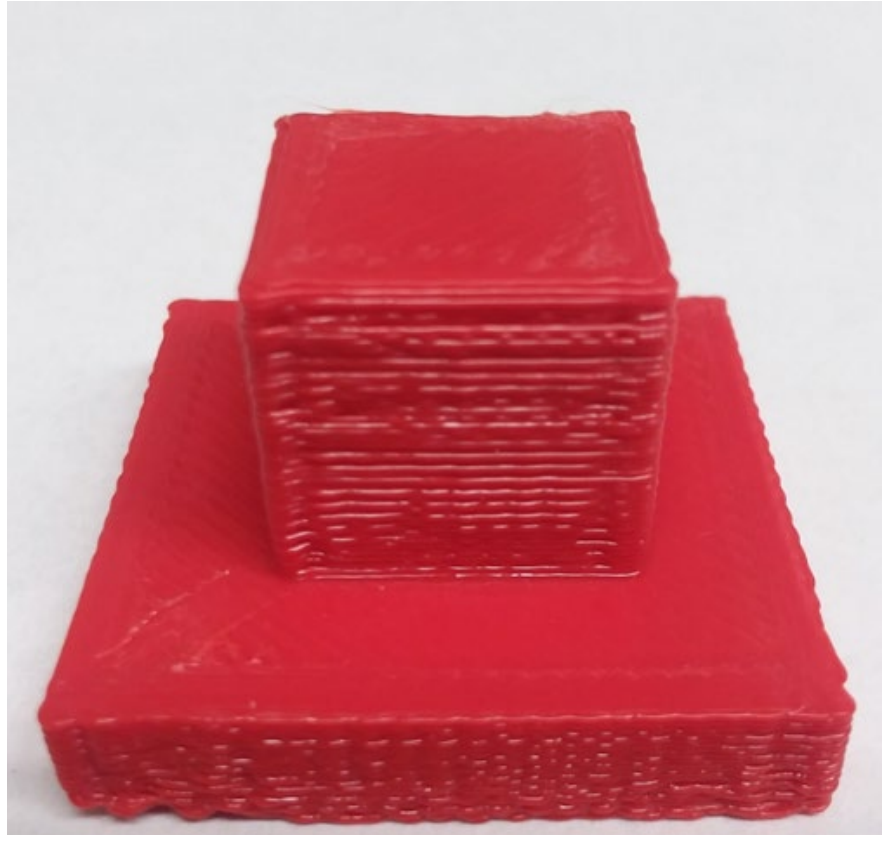

*Figure 11. Printed part with learned optimal parameter settings in both geometries.*

## 7. Conclusions and Future Work

This paper proposed an online learning-based method, namely, Continual G-Learning, to mitigate the new defects in AM process with limited samples. The proposed method addresses the challenge of limited samples in AM process by transferring offline and online prior knowledge into the current AM process. The proposed method develops an algorithm for learning the optimal defect mitigation strategies when utilizing both knowledge sources. Both numerical and real-world case studies show the effectiveness of the proposed method. In the real-world case study, the proposed method learned optimal process parameter adjustments for a printed part with two different geometries in the fewest number of actions (17 actions) compared with two benchmark methods, which need 26 and 33 actions, respectively. The results demonstrate that this proposed method significantly improves the online defect mitigation in the AM process.

The future work is focused on two directions. One direction is to investigate a decision rule to determine whether the transferred knowledge has positive or negative effects on the target process. It prevents negative knowledge transfer that can hinder the learning process. The other direction is to apply the proposed method to the multi-material AM process. Multi-material AM has various processing conditions by varying the composition and types of materials used in printing [44, 45]. Therefore, it demands a learning-based method to deal with new defects in various process conditions.

## Appendix

### A. Proof of Lemma 1

**Lemma 1.** For any policy $\pi$, the operator $B^{\pi}[CG(s,a)]$ is a contraction under the supremum norm over $s$, $a$. That is, for any $CG1(s,a)$ and $CG2(s,a)$, it follows

$$\|B^{\pi}[CG1(s,a)]\ - B^{\pi}[CG2(s,a)]\|_{\infty} \leq \gamma\|CG1(s,a) - CG2(s,a)\|_{\infty}.$$

*Proof.*

$$\begin{aligned}
&\|B^{\pi}[CG1(s,a)]\ - B^{\pi}[CG2(s,a)]\|_{\infty} \\
&= \max_{(s,a)} |k^{\pi}(s,a) + \gamma \sum_{s',a'} p(s'|s,a)\pi(a'|s')CG1(s',a') \\
&-k^{\pi}(s,a) - \gamma \sum_{s',a'} p(s'|s,a)\pi(a'|s')CG2(s',a')| \\
&= \max_{(s,a)} \left|\gamma \sum_{s',a'} p(s'|s,a)\pi(a'|s')[CG1(s',a') - CG2(s',a')]\right| \\
&\leq \gamma \max_{(s,a)} \sum_{s'a'} p(s'|s,a)\pi(a'|s')|[CG1(s',a') - CG2(s',a')]| \\
&\leq \gamma \max_{(s,a)} \sum_{s',a'} p(s'|s,a)\pi(a'|s') \max_{(s'a')}|CG1(s',a') - CG2(s',a')| \\
&= \gamma \max_{s',a'}|CG1(s',a') - CG2(s',a')| \\
&= \gamma\|CG1(s,a) - CG2(s,a)\|_{\infty}
\end{aligned}$$

where the first equality is based on the definition of the operator in Eq. (13), the first inequality is due to the triangle inequality, the second inequality is based on the property of the max operator, and the third equality is because of the property of p.m.f. function.

### B. Proof of Theorem 1

**Theorem 1.** Supposed that $0 < \rho_{min} \leq \rho_1(a|s), \rho_2(a|s) \leq \rho_{max} < 1$ for all $(s,a)$ and $\alpha_t(s_t,a_t) = n_t(s_t,a_t)^{-w}$ for $w \in (0.5, 1]$. Then, three assumptions in Theorem 2 in [34] are satisfied where $\gamma$ is the discount factor and $K = \max\left\{K' + \max_{s\in S, a\in A} CG^*(s,a)^2, 64\gamma^2\right\}$. $K'$ is defined as $2\mathbb{E}[\,R(s,a,s') - CG^*(s,a)]^2 + 4\gamma^2\left(\frac{\beta_1\beta_2}{\beta_1+\beta_2}\right)^2[2(\log|A|)^2 + 4(\log\rho_{min})^2 + 4(\log\rho_{max})^2]$. Therefore,

$$\lim_{t\to\infty} \Delta_t(s,a) \xrightarrow{w.p.1} 0.$$

According to Theorem 2 in [34], it is sufficient to prove that a random iterative process convergence of a random iterative process $\Delta_{t+1}(s_t, a_t)$ converges to zero $w.p.1$ under the following assumptions:

1. $0 \le \alpha_t(s,a) \le 1, \sum_t \alpha_t(s,a) = \infty$ and $\sum_t \alpha_t(s,a)^2 < \infty, \forall s \in S, a \in A$;
2. $\|E[F_t(s,a)|U_t]\|_\infty \le \gamma\|\Delta_t\|_\infty$, with $\gamma < 1$
3. $\text{var}[F_t(s,a)|U_t] \le K(1 + \|\Delta_t\|_\infty^2)$, for $K > 0$,

where $U_t = \{\Delta_t, \Delta_{t-1}, \dots, F_{t-1}, \dots, \alpha_{t-1}, \dots\}$ stands for the past at step $t$. The $\|\cdot\|_\infty$ refers to supremum norm and $K$ is some constant.

*Proof for Assumption 1*:

Learning rate $\alpha_t(s_t, a_t)$ is defined as $n_t(s_t, a_t)^{-w}$, $w \in (0.5, 1]$. It satisfies the assumption 1 [46].

*Proof for Assumption 2*:

Let

$$B^*[CG(s,a)] = \min_\pi B^\pi[CG(s,a)],$$

where the optimum is achieved at

$$\pi(a|s) = \frac{\rho_1(a|s)^{\frac{\beta_2}{\beta_1+\beta_2}}\rho_2(a|s)^{\frac{\beta_1}{\beta_1+\beta_2}}e^{-CG(s,a)\frac{\beta_1\beta_2}{\beta_1+\beta_2}}}{\sum_a \rho_1(a|s)^{\frac{\beta_2}{\beta_1+\beta_2}}\rho_2(a|s)^{\frac{\beta_1}{\beta_1+\beta_2}}e^{-CG(s,a)\frac{\beta_1\beta_2}{\beta_1+\beta_2}}}. \quad (14)$$

$$\begin{aligned}
&\mathbb{E}[F_t(s,a)|U_t] \\
&= \textstyle\sum_{s'\in S} p(s'|s,a)[R(s,a,s') - CG^*(s,a) - \\
&\qquad \gamma\frac{\beta_1+\beta_2}{\beta_1\beta_2}\log\left(\textstyle\sum_{a'} \rho_1(a'|s')^{\frac{\beta_2}{\beta_1+\beta_2}}\rho_2(a'|s')^{\frac{\beta_1}{\beta_1+\beta_2}}e^{-CG_t(s',a')\frac{\beta_1\beta_2}{\beta_1+\beta_2}}\right)] \\
&= B^*\big(CG_t(s,a)\big) - CG^*(s,a)
\end{aligned}$$

where the second equality is obtained by plugging Eq. (14) into Eq. (13). Therefore,

$$\begin{aligned}
\|\mathbb{E}[F_t(s,a)|U_t]\|_\infty &= \|B^*(CG_t(s,a)) - CG^*(s,a)\|_\infty \\
&= \|B^*(CG_t(s,a)) - B^*(CG^*(s,a))\|_\infty \\
&\le \gamma\|CG_t(s,a) - CG^*(s,a)\|_\infty = \gamma\|\Delta_t\|_\infty,
\end{aligned}$$

where the second equality comes from the fact that operator $B^*$ has contraction property based on Lemma 1 and monotonicity property over $CG_t(s,a)$. Both properties guarantees that applying the operator $B^*$ converges to unique optimal fixed point [47]. First inequality is based on Lemma 1.

*Proof for Assumption 3*:

Assuming that $0 < \rho_{min} \leq \rho_1(a|s), \rho_2(a|s) \leq \rho_{max} < 1$, for all $(s, a)$.

$$
\begin{aligned}
&\mathrm{var}[F_t(s,a)|U_t] \\
&\leq \mathbb{E}[F_t(s,a)]^2 \\
&= \mathbb{E}\left[\left(R\left(s,a,s^{'}\right) - CG^*(s,a) - \gamma\frac{\beta_1+\beta_2}{\beta_1\beta_2}\log\left(\sum_{a'}\rho_1(a'|s')^{\frac{\beta_2}{\beta_1+\beta_2}}\rho_2(a'|s')^{\frac{\beta_1}{\beta_1+\beta_2}}e^{-CG_t(s',a')\frac{\beta_1\beta_2}{\beta_1+\beta_2}}\right)\right)^2\right] \\
&\leq 2\mathbb{E}\left[R\left(s,a,s^{'}\right) - CG^*(s,a)\right]^2 \\
&\qquad + 2\gamma^2\frac{(\beta_1+\beta_2)^2}{(\beta_1\beta_2)^2}\mathbb{E}\left(\log\left(\sum_{a'}\rho_1(a'|s')^{\frac{\beta_2}{\beta_1+\beta_2}}\rho_2(a'|s')^{\frac{\beta_1}{\beta_1+\beta_2}}e^{-CG_t(s',a')\frac{\beta_1\beta_2}{\beta_1+\beta_2}}\right)\right)^2
\end{aligned}
$$

where the first inequality is due to the definition of variance and the second inequality is based on Cauchy–Schwarz inequality. Next, we will estimate the upper bound for the term in right hand side of above inequality.

$$
\begin{aligned}
&\mathbb{E}\left(\log\left(\sum_{a'}\rho_1(a'|s')^{\frac{\beta_2}{\beta_1+\beta_2}}\rho_2(a'|s')^{\frac{\beta_1}{\beta_1+\beta_2}}e^{-CG_t(s',a')\frac{\beta_1\beta_2}{\beta_1+\beta_2}}\right)\right)^2 \\
&\leq \max_{s'\in S}\left(\log\left(\sum_{a'}\rho_1(a'|s')^{\frac{\beta_2}{\beta_1+\beta_2}}\rho_2(a'|s')^{\frac{\beta_1}{\beta_1+\beta_2}}e^{-CG_t(s',a')\frac{\beta_1\beta_2}{\beta_1+\beta_2}}\right)\right)^2 \\
&\leq \max_{s'\in S}\left(\log\left[|A|\min_{a'\in A}\rho_1(a'|s')^{\frac{\beta_2}{\beta_1+\beta_2}}\rho_2(a'|s')^{\frac{\beta_1}{\beta_1+\beta_2}}e^{-CG_t(s',a')\frac{\beta_1\beta_2}{\beta_1+\beta_2}}\right]\right)^2 \\
&\qquad + \max_{s'\in S}\left(\log\left[|A|\max_{a'\in A}\rho_1(a'|s')^{\frac{\beta_2}{\beta_1+\beta_2}}\rho_2(a'|s')^{\frac{\beta_1}{\beta_1+\beta_2}}e^{-CG_t(s',a')\frac{\beta_1\beta_2}{\beta_1+\beta_2}}\right]\right)^2
\end{aligned}
\tag{15}
$$

where the first inequality is based on the property of max operator and the second inequality is derived by considering the range of the value in the square sign. Next, we will try to bound the first term of the last inequality in Eq. (15).

$$
\begin{aligned}
&\left(\log\left[|A|\min_{a'\in A}\rho_1(a'|s')^{\frac{\beta_2}{\beta_1+\beta_2}}\rho_2(a'|s')^{\frac{\beta_1}{\beta_1+\beta_2}}e^{-CG_t(s',a')\frac{\beta_1\beta_2}{\beta_1+\beta_2}}\right]\right)^2 \\
&\leq 2(\log|A|)^2 + 2\left(\log\min_{a'\in A}\rho_1(a'|s')^{\frac{\beta_2}{\beta_1+\beta_2}}\rho_2(a'|s')^{\frac{\beta_1}{\beta_1+\beta_2}}e^{-CG_t(s',a')\frac{\beta_1\beta_2}{\beta_1+\beta_2}}\right)^2
\end{aligned}
$$

$$\leq 2(\log|A|)^2 + 2\left(\log \min_{a'\in A} \rho_{min}{}^{\frac{\beta_2}{\beta_1+\beta_2}} \rho_{min}{}^{\frac{\beta_1}{\beta_1+\beta_2}} e^{-CG_t(s',a')\frac{\beta_1\beta_2}{\beta_1+\beta_2}}\right)^2$$

$$+ 2\left(\log \min_{a'\in A} \rho_{max}{}^{\frac{\beta_2}{\beta_1+\beta_2}} \rho_{max}{}^{\frac{\beta_1}{\beta_1+\beta_2}} e^{-CG_t(s',a')\frac{\beta_1\beta_2}{\beta_1+\beta_2}}\right)^2$$

$$= 2(\log|A|)^2 + 2\left(\log \rho_{min} + \log \min_{a'\in A} e^{-CG_t(s',a')\frac{\beta_1\beta_2}{\beta_1+\beta_2}}\right)^2$$

$$+ 2\left(\log \rho_{max} + \log \min_{a'\in A} e^{-CG_t(s',a')\frac{\beta_1\beta_2}{\beta_1+\beta_2}}\right)^2$$

$$\leq 2(\log|A|)^2 + 4(\log \rho_{min})^2 + 4(\log \rho_{max})^2 + 8\left(\frac{\beta_1\beta_2}{\beta_1+\beta_2}\right)^2 \left(\max_{a'\in A} CG_t(s',a')\right)^2$$

$$\leq 2(\log|A|)^2 + 4(\log \rho_{min})^2 + 4(\log \rho_{max})^2 + 8\left(\frac{\beta_1\beta_2}{\beta_1+\beta_2}\right)^2 \max_{a'\in A} CG_t(s',a')^2$$

where the first inequality comes from Cauchy–Schwarz inequality, the second inequality is derived by considering the range of $\rho_1, \rho_2$, and the third inequality is due to the Cauchy–Schwarz inequality. Similarly, for the second term of the last inequality in Eq. (15), we have

$$\left(\log\left[|A| \max_{a'\in A} \rho_1(a'|s')^{\frac{\beta_2}{\beta_1+\beta_2}} \rho_2(a'|s')^{\frac{\beta_1}{\beta_1+\beta_2}} e^{-CG_t(s',a')\frac{\beta_1\beta_2}{\beta_1+\beta_2}}\right]\right)^2$$

$$\leq 2(\log|A|)^2 + 4(\log \rho_{min})^2 + 4(\log \rho_{max})^2 + 8\left(\frac{\beta_1\beta_2}{\beta_1+\beta_2}\right)^2 \left(\min_{a'\in A} CG_t(s',a')\right)^2$$

$$\leq 2(\log|A|)^2 + 4(\log \rho_{min})^2 + 4(\log \rho_{max})^2 + 8\left(\frac{\beta_1\beta_2}{\beta_1+\beta_2}\right)^2 \max_{a'\in A} CG_t(s',a')^2$$

Therefore,

$\mathrm{var}[F_{\mathrm{t}}(s,a)|U_t]$

$$\leq 2\mathbb{E}\left[R\left(s,a,s^{'}\right) - CG^*(s,a)\right]^2 + 4\gamma^2\left(\frac{\beta_1+\beta_2}{\beta_1\beta_2}\right)^2$$

$$\times \max_{s'\in S}\left[2(\log|A|)^2 + 4(\log \rho_{min})^2 + 4(\log \rho_{max})^2 + 8\left(\frac{\beta_1\beta_2}{\beta_1+\beta_2}\right)^2 \max_{a'\in A} CG(s',a')^2\right]$$

$$= K' + 32\gamma^2 \max_{s'\in S, a'\in A} CG(s',a')^2$$

$$\leq K' + 64\gamma^2\left(\max_{s'\in S, a'\in A} (CG(s',a') - CG^*(s',a'))^2 + CG^*(s',a')^2\right)$$

$$\leq K' + 64\gamma^2 \max_{s\in S, a\in A} CG^*(s,a)^2 + 64\gamma^2 \max_{s'\in S, a'\in A} (CG_t(s',a') - CG^*(s',a'))^2$$

$$\leq K(1 + \|\Delta_{\mathrm{t}}\|_\infty^2)$$

where $K' = 2\mathbb{E}\left[ C(s, a, s') - CG^*(s, a) \right]^2 + 4\gamma^2 \left(\frac{\beta_1\beta_2}{\beta_1+\beta_2}\right)^2 [2(\log|A|)^2 + 4(\log \rho_{min})^2 + 4(\log \rho_{max})^2]$ and $K = \max\{K' + \max_{s\in S, a\in A} CG^*(s,a)^2, 64\gamma^2\}$. The second inequality comes from Cauchy–Schwarz inequality.

Therefore, the three assumptions in Theorem 2 are satisfied so that the proof of Theorem 1 is finished. For the generalization in terms of number of prior polices, we assume that there exist $M$ prior polices. The optimal policy can be achieved as follows:

$$\pi_{CG} = \frac{H}{\sum_a H}, \tag{16}$$

where

$$H = \exp\left(\frac{1}{\sum_{i=1}^{M} \beta_i^{-1}} \left(\Sigma_{i=1}^{M} \frac{1}{\beta_i} \log \rho_i - CG(s, a)\right)\right).$$

All the proof that we provided in the appendix can be directly applied to $M$ prior polices case with the same conclusion by plugging Eq. (16) into $\pi(a|s)$ in Eq. (13).

**C. Prior knowledge in Secs. 5 and 6**

Figure 4 in Sec. 5 shows the grid world with prior policies of Cases (a) and (b) in the numerical case study. Prior policies vary from state to state. When the state ($s$) has a single arrow, the prior policy in the direction of the arrow (e.g., right) is 0.9 and the remaining three directions have the same probability of prior policy (i.e., $\rho(right|s) = 0.9$, $\rho(left|s) = \rho(up|s) = \rho(down|s) = 0.03$). When the state ($s$) has two arrows, the prior policy in both directions (i.e., up, right) is 0.4 respectively and the remaining two directions have the equal probability (i.e., $\rho(up|s) = \rho(right|s) = 0.4$, $\rho(down|s) = \rho(left|s) = 0.1$). If the state (s) has the random prior policy, it means the probability of selecting all directions are the same with 0.25 (i.e., $\rho(up|s) = \rho(down|s) = \rho(right|s) = \rho(left|s) = 0.25$).

Table 5 in Sec. 6 shows which prior knowledge is used in each geometry in the real-world case study. The total number of actions in each state is different between geometry 1 (4) and 2 (6) since the cooling fan is excluded from the process parameters in geometry 1. States in each geometry can be expressed as the level of process parameters in the order of the flow rate multiplier, printing speed, and cooling fan specified in Table 3.

In geometry1 in experiment 1, offline knowledge that encourages to adjust flow rate multiplier as 1.0 is used. Therefore, in the states with levels of (1,1) and (1,2), 0.9 is set as the prior probability for the corresponding action, and 0.03 is equally provided to the remaining three actions. The states with the remaining levels in offline prior policy at geometry 1 have the random policy that has the same probability in four actions. In geometry 2 in experiment 1, online knowledge that is the optimal policy learned from geometry 1 is used. The optimal policy in geometry 1 is to adjust the flow rate multiplier as 1.0 at the state level (1,1,1) and printing speed to 2500 mm/min at the state level (2,1,1). The prior probability of the corresponding action is 0.9 and the probability of the remaining five actions is equal to 0.02. The states with the remaining levels in online prior policy at geometry 2 have the random policy that has the same probability in six actions. In geometry 2 in experiment 2, offline knowledge that recommends adjusting flow rate multiplier as 1.0 is utilized. Therefore, 0.9 is provided to the probability to set flow rate multiplier as 1.0, and 0.02 is provided to the each of remaining five actions in the states with levels of (1,1,1) and (1,2,1). The states with the remaining levels in offline knowledge at geometry 2 have the random policy that has the same probability in six actions.